\documentclass[11pt]{article}

\usepackage[preprint]{acl}

\usepackage{times}
\usepackage{latexsym}

\usepackage[T1]{fontenc}

\usepackage[utf8]{inputenc}

\usepackage{microtype}
\usepackage{booktabs}
\usepackage{inconsolata}

\usepackage{graphicx}
\usepackage{subcaption}
\usepackage{amssymb}
\usepackage{amsmath}
\usepackage{multirow}
\usepackage{xcolor}

\let\oldcaption\caption
\renewcommand{\caption}[1]{\oldcaption*{#1}} 

\title{AIM-Bench: Evaluating Decision-making Biases of Agentic LLM as Inventory Manager}

\author{Xuhua Zhao$^1$, Yuxuan Xie$^1$, Caihua Chen$^1$, Yuxiang Sun$^2*$\\
  $^1$School of Management and Engineering, Nanjing University \\
  $^2$School of Robotics and Automation, Nanjing University (Suzhou Campus‌) \\
  \texttt{sunyuxiang@nju.edu.cn} \\
}

\begin{document}
\maketitle
\begin{abstract}
Recent advances in mathematical reasoning and the long-term planning capabilities of large language models (LLMs) have precipitated the development of agents, which are being increasingly leveraged in business operations processes. Decision models to optimize inventory levels are one of the core elements of operations management. However, the capabilities of the LLM agent in making inventory decisions in uncertain contexts, as well as the decision-making biases (e.g. framing effect, etc.) of the agent, remain largely unexplored. This prompts concerns regarding the capacity of LLM agents to effectively address real-world problems, as well as the potential implications of biases that may be present. To address this gap, we introduce AIM-Bench, a novel benchmark designed to assess the decision-making behaviour of LLM agents in uncertain supply chain management scenarios through a diverse series of inventory replenishment experiments. 
Our results reveal that different LLMs typically exhibit varying degrees of decision bias that are similar to those observed in human beings. 
In addition, we explored strategies to mitigate the pull-to-centre effect and the bullwhip effect, namely cognitive reflection and implementation of information sharing. These findings underscore the need for careful consideration of the potential biases in deploying LLMs in Inventory decision-making scenarios. We hope that these insights will pave the way for mitigating human decision bias and developing human-centred decision support systems for supply chains. 
\end{abstract}

\section{Introduction}

In recent years\footnote{This is a preprint.}, rapid advances in the mathematical reasoning\cite{wang2025mvmathevaluatingmultimodalmath} and code generation capabilities\cite{zhang2025artifactsbenchbridgingvisualinteractivegap,ye2025uncovering} of LLMs have enabled the LLM agent to revolutionize various fields. Furthermore, the utilization of LLM agents in business operations is gaining prominence\cite{Huang2022FinBERT,Liu2025Mitigating}. Decision models to optimize inventory levels are one of the core elements of operations management. Hence, research into the use of LLM agents for inventory management has significant and realistic value. Forbes highlights the specific ways in which LLMs could be used in operational tasks: “with ChatGPT, retailers can manage inventory levels by analyzing sales data and predicting demand. This can help retailers avoid overstocking or running out of products, so they reduce costs and keep customers happier.”\cite{fbs2023}.\footnote{$*$Corresponding author. }

\begin{figure}
    \centering
    \includegraphics[width=1\linewidth]{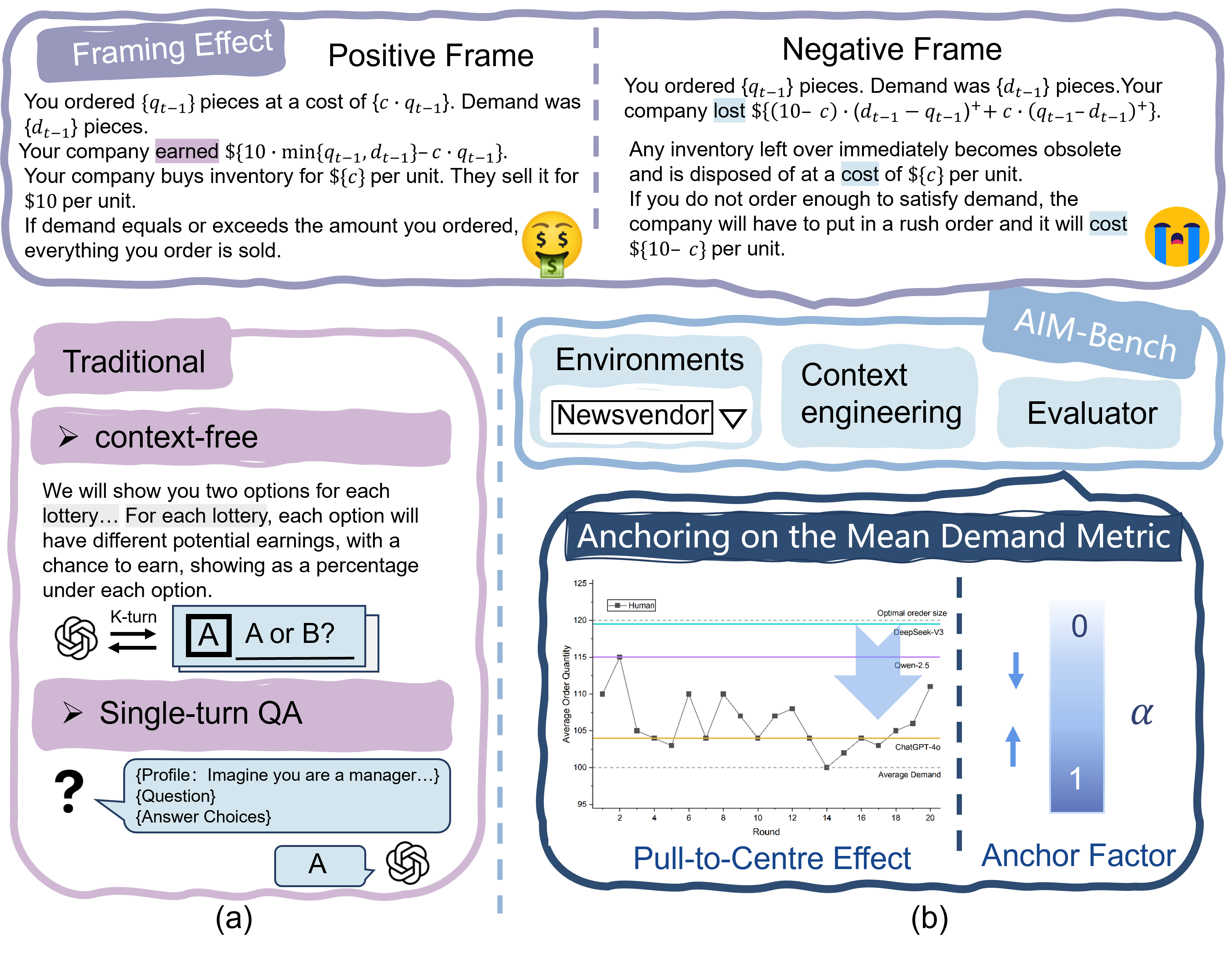}
    \caption{An overview of the evaluation of Pull-to-Centre Effect in the Newsvendor problem, where (a) is the traditional behavioural evaluation and (b) is the interaction protocol of AIM-Bench}
    \label{fig:1}
\end{figure}

But unlike Forbes's way, this paper will study the capability of LLMs-enabled inventory agents to make end-to-end replenishment decisions. In the extant literature, Quan\cite{quan2025invagentlargelanguagemodel} et al. initiated investigation into the performance of LLM agent replenishment decision-making. However, this investigation was confined to the Beer Game (BG)\cite{Forrester1958}. In the context of other replenishment decision problems, including the newsvendor problem\cite{Arrow1951}, the multi-period inventory replenishment problem, and the dual-source inventory management problem, there remains a significant gap in the existing literature. Furthermore, it has been demonstrated that individuals frequently exhibit behaviours that deviate from rationality when confronted with decisions in uncertain circumstances.\cite{Tanaka2010Risk} One of the most well-known and highly disruptive results of these deviations from optimal behavior is the bullwhip effect (BWE)\cite{Lee1997The}, which describes the phenomenon of order variance amplification in supply chains. 
Do LLMs suffer from decision bias in inventory replenishment problems as humans do? Since these factors have sparked widespread concern\cite{cui2025large,jia2024decision,chen2025manager},  a comprehensive evaluation of LLM agents is crucial for the progression of this emerging field.

To address these issues, we introduce AIM-Bench, a benchmark designed for inventory LLM agents, for detailed decision-making performance and bias assessment. AIM-Bench encompasses a diverse set of 4 different behavioural assessment tasks and 5 demonstration environments, ranging from single-agent, single-period newsvendor problem (NVP) to multi-agent, multi-period complex supply chain networks (SCN). Notably, each environment, whether newly created or adapted from a pre-existing environment, is subject to consideration of its one or more sources of uncertainty in the context of the supply chain. Uncertainty sources include stochastic demand, stochastic VLTs, and uncertain behaviour of supply chain partners.

Using the five proposed environments, we evaluate the inventory and ordering decision-making capabilities of various popular open-source and closed-source LLMs. We use real-world metrics such as inventory management costs, stockout rates and turnover rates as the core criteria for assessing agent effectiveness. In addition to the outcome metrics, the decision-making process also provides rich evaluative potential. In multi-period replenishment (MPR), we employ a fine-grained metric to track the gap between agent replenishment actions and the optimal action, which gives us a thorough understanding of the resulting trajectories. Furthermore, in order to investigate the decision-making biases of LLM agents, we use the NVP and the beer game to investigate four common human biases in the decision-making process, i.e., prospect theory, mean anchoring, demand chasing, and BWE. We found that only pointed closed-source LLMs such as ChatGPT-4.1 and Gemini-2.5F-L were effective in overcoming the decision bias on the newsvendor model, but their experiments in overcoming the bullwhip effect were disappointing. Another interesting finding is that, in contrast to existing literature on framing effects, LLMs found no evidence of risk reversal in the context of NVP. 

In this paper, we make the following contributions:

\begin{itemize}
    \item We introduce AIM-Bench, the first benchmark specifically designed to evaluate LLM agents' inventory decision-making under multi-source uncertainties (stochastic demand, stochastic lead times, and partner behavior uncertainty) across 5 diverse supply chain environments.
    \item We conduct the evaluation of 5 open/closed-source LLMs on end-to-end replenishment tasks through real-world metrics and fine-grained process metrics.
    \item We develop a methodology to detect and quantify human-like decision biases (prospect theory, mean anchoring, demand chasing, BWE) in LLMs, revealing significant inventory and ordering decision-making bias in most models. In addition, we explored strategies to mitigate the pull-to-centre effect and the BWE, namely cognitive reflection and implementation of information sharing.
\end{itemize}
\begin{figure*}
    \centering
    \includegraphics[width=1\linewidth]{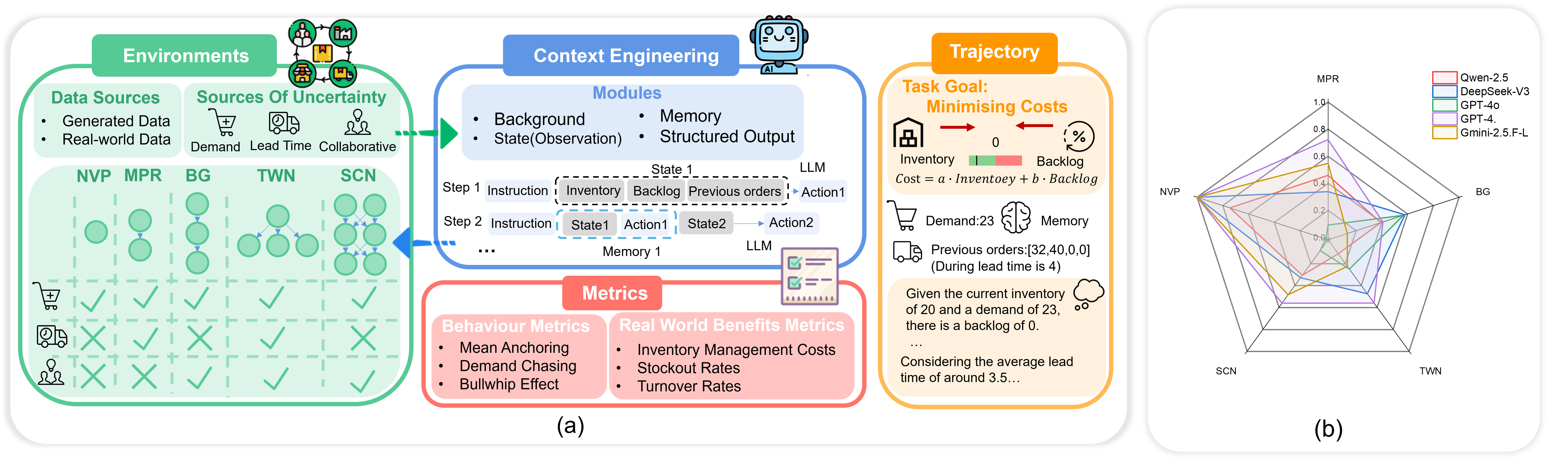}
    \caption{ (a) The illustrative overview of AIM-Bench. AIM-Bench consists of five environments containing multiple uncertainties. We categorised the environments according to the uncertainties. (b) Management scores for different LLMs across environments.}
    \label{fig:2}
\end{figure*}

\section{Related Works }

In this section, we first briefly review existing methods for NVP, BG, MPR and other inventory management tasks. After that, we will discuss some existing literature on social science approaches to decision bias in LLMs.

\textbf{AI For Inventory Management }The NVP is a classic inventory problem introduced in 1951\cite{Arrow1951}. Schweitzer et al.\cite{schweitzer2000decision}were among the first to analyze human decision makers in the newsvendor setting and identify the pull‐to‐center effect. The multiperiod inventory management problem has been studied in the decades since Kaplan\cite{kaplan1970dynamic}.  Qi et al. \cite{Qi2022} construct near-optimal ordering policies with a one-step end-to-end deep learning framework for inventory problems with non-stationary demands and stochastic lead times.

BG was first introduced in the late 1950s by \cite{Forrester1958}. It has grown over the decades to become a popular tool for illustrating some of the fundamental challenges associated with Inventory and ordering decisions supply chain management.  Some researchers \cite{Liu2025,oroojlooyjadid2022deep}have developed replenishment policies using deep reinforcement learning (DRL) algorithms to mitigate demand overestimation in serial supply chain networks (e.g. BG) and the resulting BWE. 

Several benchmarks and toolkits for LLM agents have been established, focusing on various tasks such as web-browsing, games, and tool use\cite{wang2025mchirc,zhuang2025pokerbench,yuan2025dmt,zhou2025characterbench,xie2025expand,zhou2025urbench}. Currently, only the BG environment has been studied for LLM-based agents\cite{quan2025invagentlargelanguagemodel}, and there is still a gap in the benchmark for decision-making in LLM replenishment. . However, there is a lack of research on more complex supply chain replenishment decisions. Our study focuses specifically on text-based environments that encompass various uncertainties, in order to evaluate the capabilities of LLM agents in replenishment decisions.

\textbf{Decision making under Uncertainty in LLMs. } 
Numerous studies in the social sciences have used behavioural theories to assess LLMs' decision-making under uncertainty. For example, one large-scale study used LLMs to replicate 156 psychology experiments from leading social science journals\cite{cui2025large}.Another study assessed the risk appetite, probability weighting and loss aversion of LLMs.\cite{jia2024decision} A key objective of this area of research is to evaluate and enhance the capacity of LLMs to substitute for human research participants in experiments, a field known as the social science of AI.\cite{dillion2023can,wang2025large}. In these experiments, LLMs showed characteristics consistent with human behavioural patterns.

However, our ultimate goal is to enhance decision support systems by investigating the biases of LLMs to improve human operational decision-making capabilities.For example, Su et al. \cite{su2023can} explore the capabilities of the GPT-4 with the classic NVP, while Chen et al.\cite{chen2025manager} study decision bias in an inventory context. Although these works have assessed LLMs decision-making abilities and biases in specific contexts, in the real world LLMs tend to make decisions as agents by attaching memories and perceptual models. We aim to assess decision-making behaviour through LLM agents interacting with the environment.

\section{AIM-Bench}

\subsection{Task Formulation }

In our inventory and ordering environments, LLM agents receive a textual description of the task context and status. And agents are prompted to output the next action as a natural language string, conditioned on their past interaction history.

We formulate these environment as a Partially Observable Markov Game (POMG)\cite{littman1994markov} , which can be defined as a tuple  $\mathcal{G} = \left( \mathcal{N}, \mathcal{S}, \{ \mathcal{A}_i \}_{i \in \mathcal{N}}, \{ \mathcal{O}_i \}_{i \in \mathcal{N}}, \mathcal{P}, \{ \mathcal{R}_i \}_{i \in \mathcal{N}}, \gamma \right)$, where $\mathcal{N} = \{1, \cdots, n\}$ is the set of agents; $\mathcal{S}$ is the state space; $\mathcal{A}_i$ and $\mathcal{O}_i$ are the action space and observation space of agent $i$, respectively; $\mathcal{P} : \mathcal{S} \times \{ \mathcal{A}_i \}_{i \in \mathcal{N}} \to \Delta(\mathcal{S})$ is the transition function; $\mathcal{R}_i : \mathcal{S} \times \{ \mathcal{A}_i \}_{i \in \mathcal{N}} \to \mathbb{R}$ is the reward function of agent $i$; and $\gamma$ is the discount factor. In order to make LLM agents simulate human-like behavioural decisions, we added the memory module \(m_{i,t}=\{o_{i,t-k},a_{i,t-k}...o_{i,t-1},a_{i,t-1}\}\), which is a sequence of actions and observations. 

\subsection{Environments }

\textbf{Newsvendor Problem}. The newsvendor model considers a decision maker who must determine the order quantity of a product for a single selling period.

\textbf{Multi-period Replenishment Decision.} We consider the multi-period inventory management problem allow for inventory to carry over from one period to the next, making them reflective of a broader range of consumer products in the marketplace. There are two types of uncertainties (stochastic demand and stochastic VLT), whereas the NVP only considers one source of uncertainty (stochastic demand). 

\textbf{Beer Game.} Game includes players who represent four echelons in a hypothetical beer supply chain (production plant, distributor, wholesaler, and retailer). Gameplay involves the players collaborating in an attempt to fulfil customer demand while minimizing costs \cite{Croson2014}. 

\textbf{Two-level Warehouse Network. } The two-level warehouse network consists of a central warehouse upstream near the manufacturer, acting as a transfer hub, and multiple mini-warehouses downstream, closer to customers. Goods are first bulk-shipped to the central warehouse and then dispatched to mini-warehouses based on demand; mini-warehouses can also order directly from the manufacturer, with direct shipments slower than central warehouse-to-mini-warehouses ones but faster than two-step routes.

\textbf{Supply Chain Network.}  Similar to the dual-source inventory management problem, the downstream level of the network has two upstream distributers, one of which is considered a normal source of supply and the other an expedited source of supply with a higher selling price and shorter lead time.

The supply chain structure and sources of uncertainty are shown in Figure \ref{fig:2}.

\section{Evaluation Protocol }

\subsection{Newsvendor Problem and Prospect theory}

The newsvendor model considers a decision maker who must determine the order quantity \( q\) of a product for a single selling period. The unit purchase cost of the product is \(c\), and the unit revenue is \(r\). The demand of the product \( D\) is stochastic with probability density function \(f ( d)\).Stocks must be ordered during the previous period. Excess stocks at the end of the day become obsolete at a cost. Shortages results in a loss of possible sales. If the demand realization \(d\) exceeds the order quantity \(q\), \(q\) units are sold, and excess demand is lost. If the demand realization \( d\) is less than the order quantity \( q\), \(y\) units are sold, and  \(q-d\) units are left over and salvaged at unit salvage value \(0\). 

The critical ratio corresponds to the profit margin applicable in a newsvendor scenario, where \(c_u\) is underage cost (demand that cannot be satisfied using inventory) and \(c_0\) is overage cost (excess inventory that cannot be sold). The unit underage cost is $c_{\text{u}} = r - c$. ; the unit overage cost is $c_{\text{o}} = c$. 
\begin{equation}
\text{Critical ratio} = \frac{c_u}{c_u + c_o}
\end{equation}

As shown in Figure \ref{fig:1}. We frame newsvendor decisions as losses vs. profits, aiming to explore the risk preference of LLM in the NVP. LLM agents were asked to make inventory purchase decisions, \(q_t\), for 20 rounds of a standard newsvendor problem. The LLM prompts consisted of two frames: a positive frame (PF) emphasizing profit and a negative frame (NF) emphasizing cost. Under the positive frame, LLM agents were essentially asked to maximize the newsvendor’s expected profits. Under the negative frame, they were instead asked to solve the equivalent problem of minimizing the drop below the expected profits under perfect information.

Two natural anchors in the newsvendor context are the mean demand and previous demand realizations. Usually, people base decisions initially on an available anchor and then use additional information to adjust the decision toward the optimal decision. The final decision is insufficiently adjusted and biased toward the initial anchor.. The degree to which the LLM agent is anchored is quantified using the following metrics:

\subsubsection{\textbf{Anchoring on the Mean Demand}}
Anchoring on the mean demand can be modeled by using an anchor factor \(\alpha\) that quantifies the degree to which people anchor. People initially anchor on the mean demand \(\mu\) and then adjust a fraction \((1-\alpha)\) of the optimal adjustment \(q^{*} - \mu\). The factor\( (1-\alpha)\) can be interpreted as an adjustment factor; we denote this factor by \(\alpha' = (1-\alpha)\) \cite{bolton2008learning}. The adjustment process can be modeled as

\begin{equation}
q = \mu + \alpha'(q^{*} - \mu).
\end{equation}

\subsubsection{\textbf{Demand Chasing Heuristic.}}
The demand chasing heuristic is another anchoring and insufficient adjustment heuristic. It states that people anchor at previous order quantities and adjust them toward prior demand realizations. And a simple measure of demand chasing is the correlation between current orders and previous demand, \(\rho(q_t, d_{t-1})\)\cite{bolton2008learning}. For i.i.d. demand, the correlation between previous and current demand is zero, where positive correlation indicates demand chasing behavior.

\subsection{Multi-period Replenishment Metrics}

We consider the multi-period inventory management problem and the details are as follows: for a single item at a single location, we consider a finite horizon of discrete periods \(1, \ldots, T\), where \(T\) is the end of the horizon. Over the \(T\) periods, there is a sequence of random demands, denoted by \(D_t\), \(\forall t = 1, \ldots, T\). Let \(I_t\) denote the inventory level at the beginning of period \(t\). The inventory level can be positive if we have inventory excess on hand or negative if we have an inventory shortage and, hence, backorders. As a result, at the end of each period, we either incur a holding cost of \(h\) for each excess unit or a stockout cost of \(b\) for each backordered unit.

We consider periodic review policies and assume that review periods are given as a sequence of dates. That is, we assume there are in total \( M \) orders from period \( 1 \) to \( T \) that are placed at \( t_m \), \( \forall m = 1, \ldots, M \). This assumption is aligned with the real-world practice, in which a fixed schedule is typically held for order placement (e.g., one can place orders on Tuesdays and Fridays). 

In this problem, we consider the stochastic VLT. That is, the \( m \)-th order placed at period \( t \) arrives at period \( t + L \), where \( L \) is a random variable that only takes positive integer values. Hence, the arrival time of orders, denoted by \( v_m = t_m + L_m \), \( \forall m = 1, \ldots, M \), are also random variables. Moreover, we assume there are no crossing over of order arrivals. 

Hereafter, \( D_t \) and \( L_m \) denote the random variables; \( d_t \) and \( l_m \) denote the realization of demand at period \( t \) and the realization of VLT of the \( m \)-th order, respectively. At each period, the system first updates the inventory level by checking if any order has arrived; then, demand occurs. Let \( a_m \) denote the order quantity for the \( m \)-th order. At the end of the period, either a holding or backorder cost occurs. The inventory level updates follow the equation

\begin{align}
    I_{t + 1} = I_t - D_t + \sum_{m = 1}^{M} a_m \mathbf{1}\{ t = t_m + L_m \}
\end{align}

Let the cost that occurred at period \( t \) be denoted by \( C_t \), and we have

\begin{align}
    \begin{split}
C_t &= h\left[I_t - D_t + \sum_{m = 1}^{M} a_m \mathbf{1}\{t = t_m + L_m\}\right]^+  \\
    &\quad + b\left[-I_t + D_t - \sum_{m = 1}^{M} a_m \mathbf{1}\{t = t_m + L_m\}\right]^+.
\end{split}
\end{align}

Here, \([\cdot]^+\) denotes \(\max\{\cdot, 0\}\).

Inspired by the end-to-end inventory management model \cite{Qi2022}, we compute the expost optimal quantity for each order by means of a dynamic programming framework, which is used as the answer to the intermediate step. And intermediate step is evaluated per period order quantity using distance from optimal order quantity.

Specifically, Qi et al. demonstrated that the optimal multi-period inventory replenishment problem described by (4) is decomposable. In addition, we can get the ex-post optimal order quantity \(a_m^{*}\) by using the following equation:

\begin{equation}
     a_m^{*} = \max\left\{ d_{[v_m, s^*]} - I_{v_m}, 0 \right\}
\end{equation}

where \(s^* = \left\lfloor \frac{h(v_{m+1} - v_m)}{h + b} \right\rfloor + v_m \) ,\(s^*\) is optimal replenishment period. 
Calculate the distance from optimal order quantity using the formula:

\begin{equation}
    d(\mathbf{a}, \mathbf{a^{*}}) = \sqrt{\sum_{i=1}^{m}(a_m - a_m^{*})^2}
\end{equation}

\subsection{Beer Game and Bullwhip Effect.}

In this problem, we consider the fixed lead time. In each period \( t \), the quantity of goods that are shipped by actor \( i \) is represented by \( a_t^i \). We use \( a_t^{M + 1} \) to represent the quantity of goods shipped by the manufacturer to actor \( M \) in period \( t \). All goods shipped by agent \( i \) will be in transit for \( L \) periods and added to actor \( i + 1 \)'s stocks in period \( t + L \). Hence, the in-transit orders for actor \( i \) at the beginning of period \( t \) can be represented by the vector \( \left( a_{t - 1}^{i + 1}, a_{t - 2}^{i + 1}, \ldots, a_{t - L}^{i + 1} \right) \). For each period \( t \), agent \( i \)'s sellable goods are made up of its remaining inventory \( I_t^i \) and just arrived orders \( a_{t - L}^{i + 1} \). 

In the BG, since each agent only has one upstream source for goods replenishment, each agent's ordering is constrained by the inventory level of its immediate upstream supplier. In TWN and SCN, there are two suppliers of downstream agent, one of which is considered a normal source of supply and the other an expedited source of supply. likewise, other than ordering from the central warehouse, each mini-warehouse can also make orders directly from the manufacturer in Two-level Warehouse Network. The BG inventory level updates follow the equation:
\begin{equation}
    I_{t + 1}^i =  I_t^i + a_{t - L}^{i + 1} - a_t^i 
\end{equation}
The overall costs of the system in period \(t\) can be expressed by

\begin{multline}
C_t^{\text{total}} = \sum_{i=1}^{M} \bigg[ h^i \left( I_t^i + a_{t-L}^{i+1} - a_t^i \right)^+  \\
+ b^i \left( -I_t^i - a_{t-L}^{i+1} + a_t^i \right)^+ \bigg]
\end{multline}

One of the most well-known and highly disruptive results of these deviations from optimal behavior is the BWE, which describesthe phenomenon that demand fluctuations increase from downstream to upstream in a supply chain\cite{Lee1997The}.

While normative behavior can lead to the BWE \cite{Lee1997The}. studies conducted in a controlled laboratory environment provide evidence that the problem persists even when researchers control for these rational causes \cite{croson2003impact,croson2006behavioral,croson2002experimental}, demonstrating that irrational decisions by managers can be a significant source of the problem.

Information sharing between different supply chain roles have been extensively studied as a way to mitigate BWE \cite{zhao2015human,sarkar2015behavioral}.

In the context of the BWE analysis in supply chain management, the mathematical formulations for the standard deviation of demand variability across different echelons are presented as follows:

The BWE is formally quantified as the ratio of demand variability between two consecutive echelons in a supply chain. Let \(D_{t}^{(1)}\) denote the demand at the downstream echelon (e.g., retailer) in period \(t\), and \(D_{t}^{(2)}\) the demand at the immediate upstream echelon (e.g., distributor). The BWE metric \(\beta\) is defined as:

\begin{align}
    \beta = \frac{\sigma_2}{\sigma_1}
\label{eq:bwe}
\end{align}

where \(\sigma_1 = \text{std}(D_t^{(1)})\) and \(\sigma_2 = \text{std}(D_t^{(2)})\) denote the standard deviations of downstream and upstream demand, respectively. A value \(\beta > 1\) indicates demand volatility amplification (BWE), \(\beta = 1\) implies no amplification, and \(\beta < 1\) suggests demand smoothing.

To further explore the ability of LLM agent to solve real-world inventory management problems, we consider three benefit metrics:inventory management cost, stockout rate and turnover rate. These metrics are used in real logistics companies to evaluate the performance of replenishment algorithms\cite{Qi2022}.

\subsection{Real-world evaluation metrics}

\textbf{Average Inventory Management Cost.} In Multi-period Replenishment Decisions problem, inventory management cost is given by equation, and cost is given by equation in BG. For more details of cost parameters and other problems inventory mamagement cost equation, please refer to Appendix B. 
\begin{align}
    \textbf{AC} = \frac{\sum_{t}^T C_t}{T},
\end{align}

\textbf{Turnover Rate.}The inventory turnover rate is calculated by dividing the average inventory level of each period\( \frac{\sum_{t}^T I_t}{T}\) by the average demand \( \frac{\sum_{t}^T D_t}{T}\), hence it is given by
\begin{align}
    \textbf{TR} = \frac{\sum_{t}^T I_t}{\sum_{t}^T D_t},
\end{align}

\textbf{Stockout Rate.}The stockout rate is defined as the percentage of period that stockout \(T_S\) occurs during the experimental period \(T\),hence it is given by

\begin{align}
    \textbf{SR} = \frac{ T_s}{ T},
\end{align}

\section{Experiments }

\subsection{Models}

We evaluate a range of popular closed-source and open-source models, including DeepSeek-V3\cite{deepseekai2025deepseekv3technicalreport}, and Gemini-2.5-flash-lite \cite{comanici2025gemini25pushingfrontier}, GPT-4.1 (2025-04-14 release) and GPT-4o (2024-11-20 release) \cite{openai2024hello},  as well as Qwen-2.5-72B\cite{qwen2025qwen25technicalreport}.

\subsection{Main Result and Findings}

\begin{figure}
    \centering
    \includegraphics[width=1\linewidth]{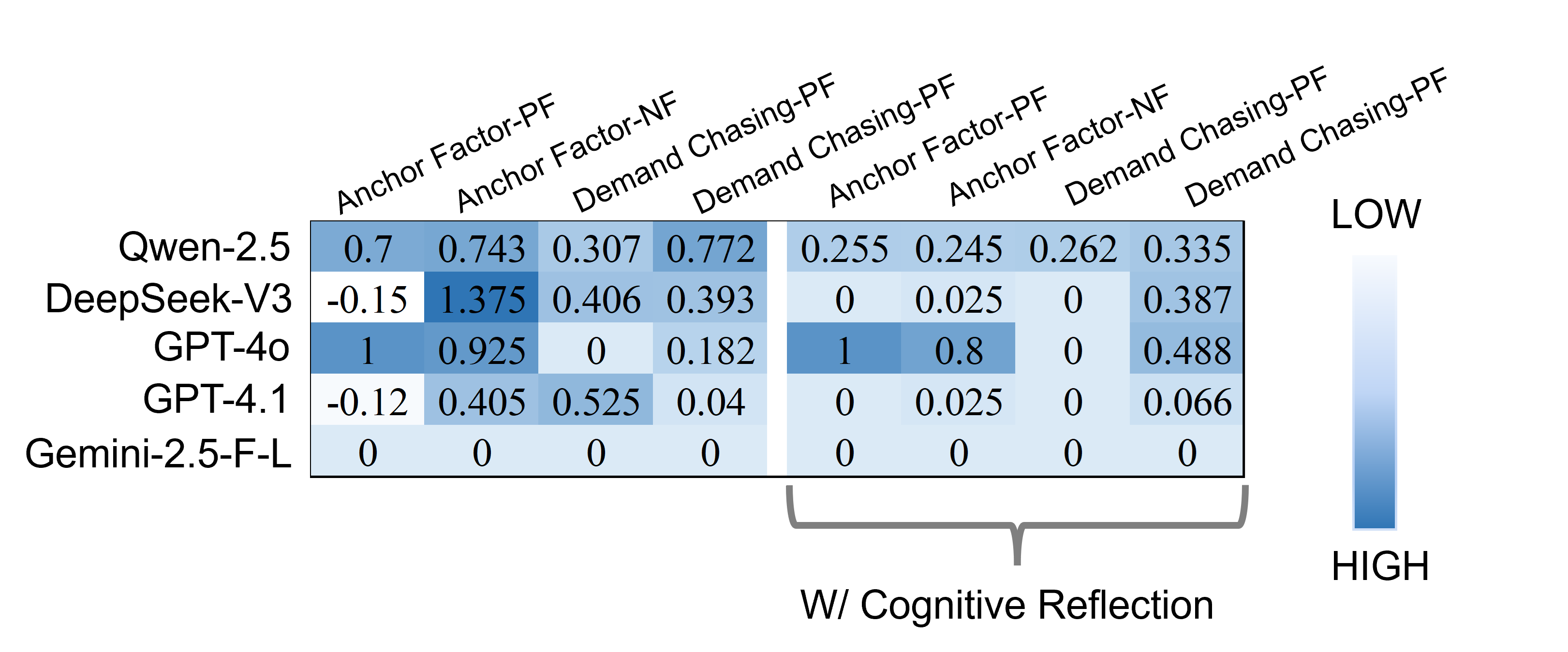}
    \caption{Result in NVP with c = \$3 in the high-margin treatment. }
    \label{fig:A}
\end{figure}

\textbf{NVP-Finding I : Most of the evaluated LLMs show a tendency to anchor on the mean demand.} Figure \ref{fig:A} shows higher anchoring factors, indicating that LLMs are susceptible to anchoring and insufficient adjustment bias. For example, GPT-4o exhibits significant demand anchoring with an \( \alpha\) of 1 and 0.925. Even state-of-the-art LLMs like GPT-4.1 and DeepSeek show substantial decision-making bias, particularly in anchoring factor (i.e., GPT-4.1: 0.405; DeepSeek-V3: 1.375). Only Gemini-2.5-flash-lite showed complete immunity to anchoring and insufficient adjustment bias. One explanation for the anchoring of the observed demand is that the model does not use logical reasoning to arrive at an answer, but rather statistical pattern matching over large amounts of text. The high-frequency association of the initial anchor point with the answer enables it to be influenced by the apparent context.

In addition, in the demand chase problem, LLMs demonstrates significantly less bias in its estimates compared with humans, often providing unbiased dominant responses.

\textbf{NVP-Finding II: Cognitive reflection can mitigate decision-making bias to some extent.} The motivation for using cognitive reflection from two core observations: i) Finding I shows that LLMs struggle to make optimal decisions. ii) Previous works suggest that adjusting LLMs’ persona can enhance specific capabilities. Bearing these insights in mind, a promising strategy for mitigating demand anchoring is to enhance the scaling capacity of LLMs in order to facilitate cognitive reflection. Specifically, we have designed a new prompt to imitate human System 2 thinking. As seen, these modifications result in lower anchoring factor. For example, alpha decreased from 0.7 to 0.255 for Qwen-2.5 in positive framework and from 0.74 to 0.245 in negative framework.

\begin{figure}[htbp]
    \centering
    
    \begin{minipage}[b]{0.49\linewidth}  
        \centering
        \includegraphics[width=0.99\linewidth]{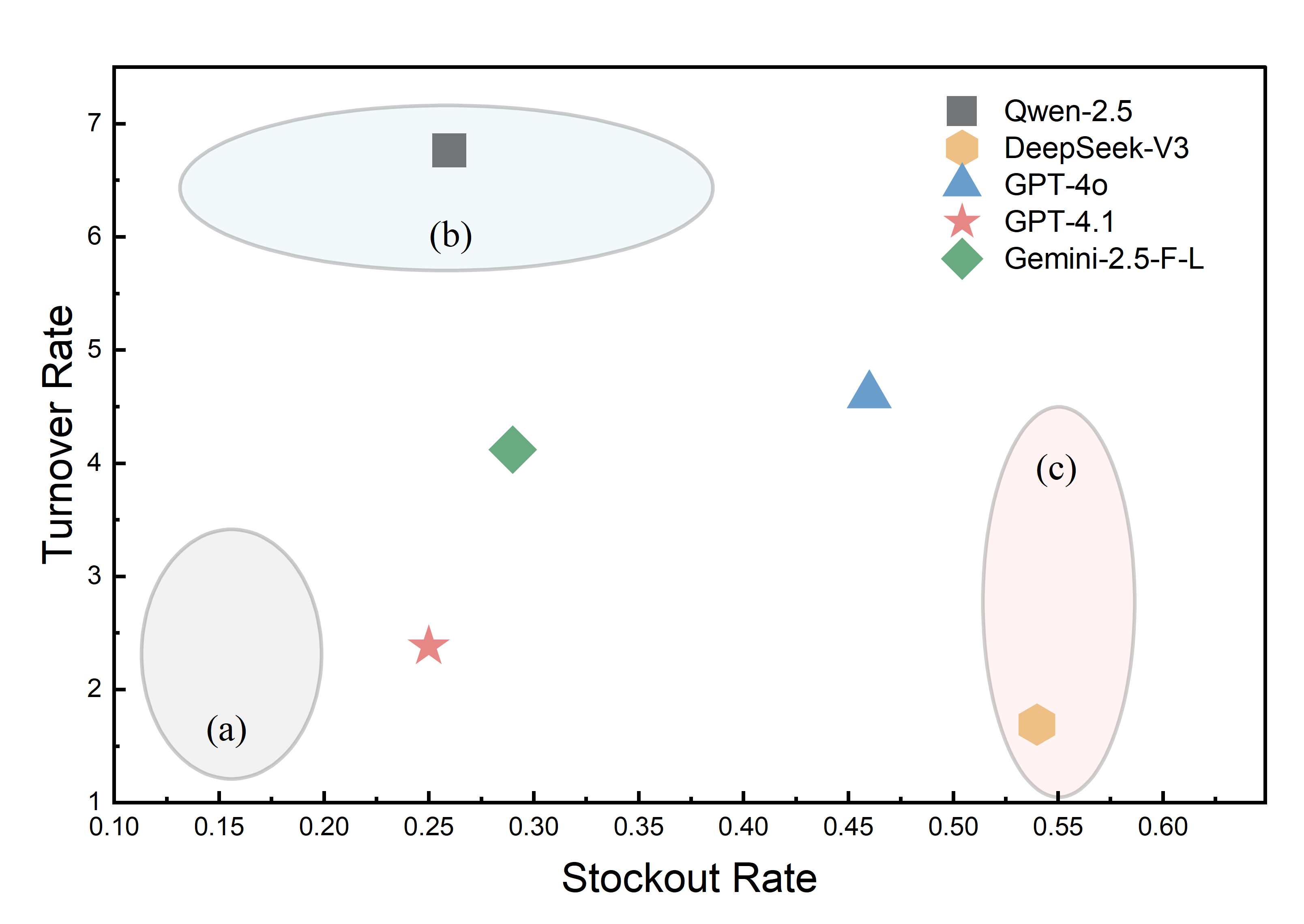}  

    \end{minipage}
    \hfill  
    \begin{minipage}[b]{0.495\linewidth}  
        \centering
        \includegraphics[width=0.99\linewidth]{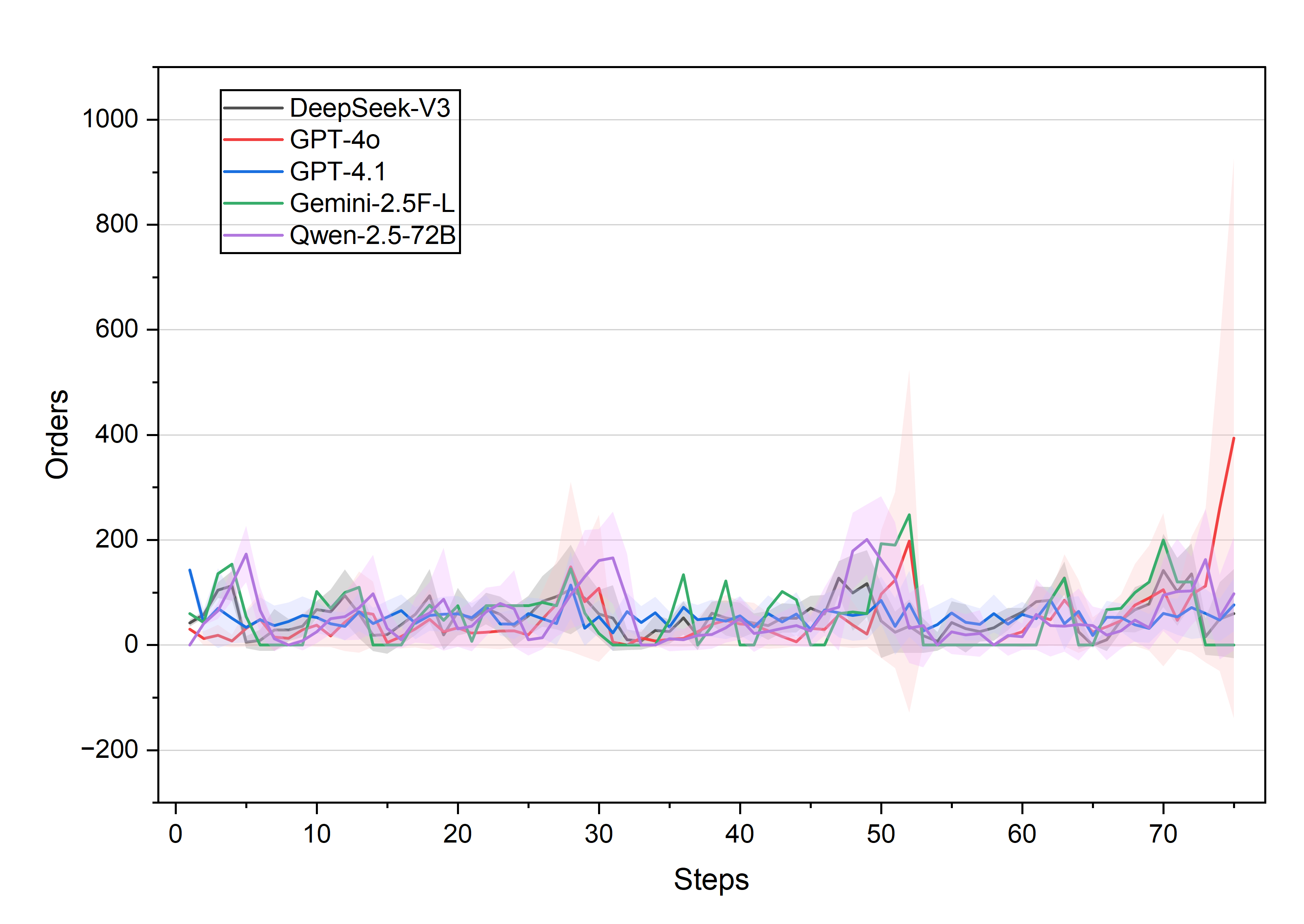}  

    \end{minipage}
    
    \caption{ Order Processes in Multi-period Inventory Decision Making with Different LLM Agents. Of these, GPT-4o and Qwen exhibit significant over-estimation of demand uncertainty, thus over-ordering inventory. }  
    \label{fig:TR-SR}  
\end{figure}

\textbf{NVP-Finding III: Behavioral theories need to be tested in context.} Previous studies have surfaced the presence of implicit risk aversion in LLMs. Ma et al. \cite{ma2023chatgpt} use the disease outbreak problem to study the decision-making framework and observe that the GPT prefers risk in case of loss and certainty in case of gain. Inspired by this, we investigate the use of framing to change newsvendors’ risk preference in order to induce them to make better ordering decisions. However, to our surprise the experimental results did not find evidence of risk reversal. Thus, we conclude that risk reversal cannot reliably be used without pretesting and that LLMs behavioral theories need to be tested in context.

\begin{figure}
    \centering
    \includegraphics[width=1\linewidth]{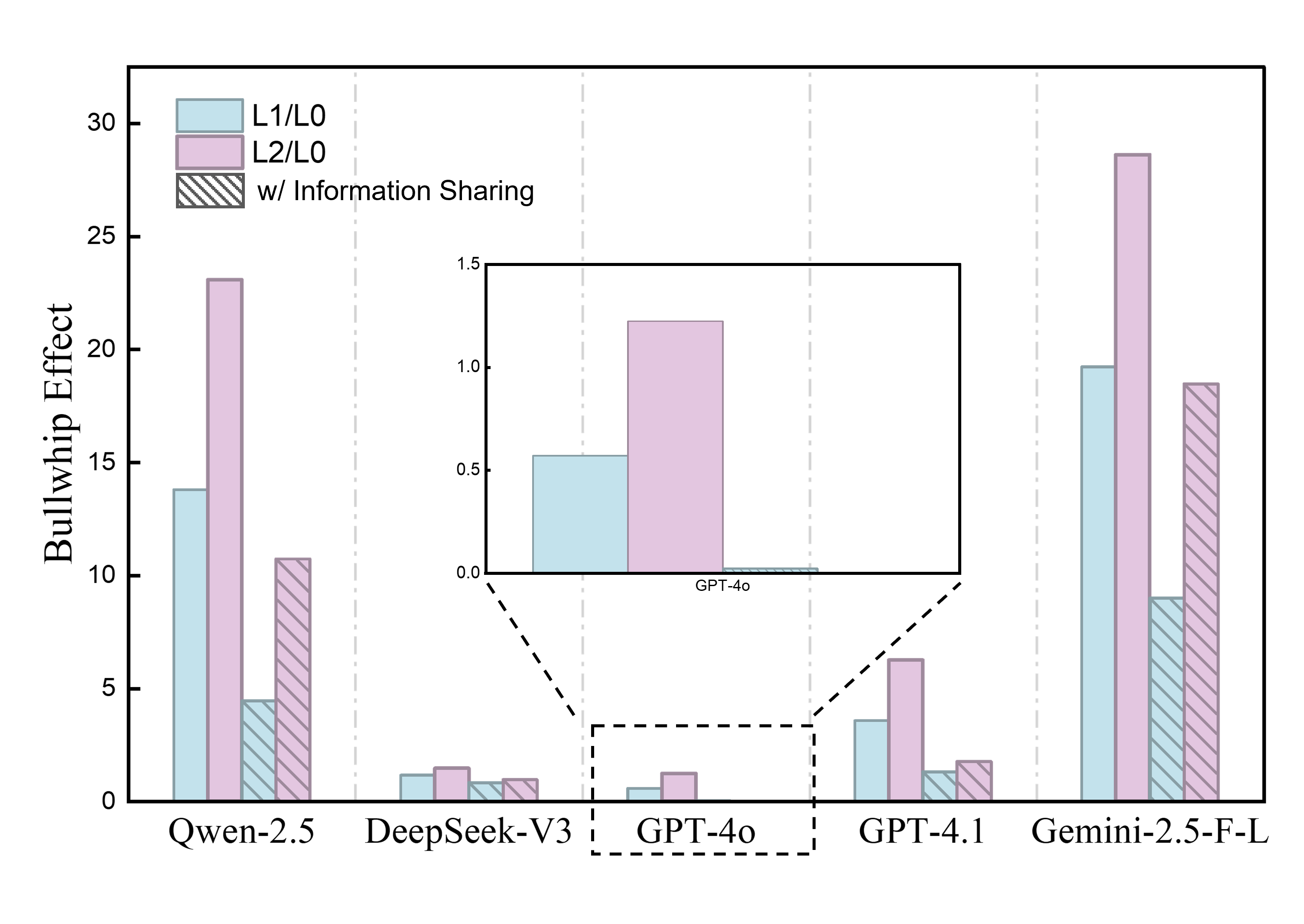}
    \caption{BWE of BG. The bullwhip effect was significantly reduced in the information sharing setting. However GPT-4o 
exhibited action chasing towards partners}
    \label{fig:is}
\end{figure}

\textbf{BG-Finding IV: All the evaluated LLMs show the BWE due to demand overestimation.} Results in Fig \ref{fig:is} indicate that the bullwhip effect is widespread. For example, Gemini-2.5-flash-lite, the best performer in the newsvendor, demonstrates the highest BWE of 19.22 and 28.61. Even state-of-the-art LLMs like GPT-4.1 show substantial demand overestimation. Similarly, information sharing is a promising approach to mitigating the bullwhip effect, inspired by human heuristic decision-making. Specifically, we design the new state space, which can acquire so information about partners as agent context, to replace the original partially observable observation space. The results show that information sharing can significantly reduce the bullwhip effect. For example, the bullwhip effect for Qwen-2.5 was reduced from 13.78 to 4.45 and from 23.07 to 10.73. However, GPT-4o shows a tendency to action chasing and conform under the information-sharing setting, which restricts the agent's ability to explore different strategies (The BWE is close to 0).

\begin{table*}[htbp] 
\centering
\setlength{\tabcolsep}{3pt} 
\renewcommand{\arraystretch}{1.2} 
\begin{tabular}{c lllc c c c c c c c c} 
\toprule 
\multirow{2}{*}{Model} &   \multicolumn{3}{c}{BG}&\multicolumn{3}{c}{MPR} & \multicolumn{3}{c}{TWN} & \multicolumn{3}{c}{SCN} \\
\cmidrule(lr){2-4} \cmidrule(lr){5-7} \cmidrule(lr){8-10}\cmidrule(lr){11-13} 
 &   Avg. C& TR&SR&Avg. C& TR& SR& Avg. C& TR& SR& Avg. C& TR& SR\\ 
\midrule 
Qwen-2.5-72B &   21.28& 0.40&1.73&649& 6.76& 0.26& 1779& 7.12& 0.33
& 1039& 8.53& 0.29
\\
DeepSeek-V3 &   14.72& 0.02&1.59&794& 1.69& 0.54& 1048& 1.45& 0.51
& 1007& 3.78& 0.47
\\
gpt-4o &   15.14& 0&2.91&1090& 4.60& 0.46& 1498& 4.28& 0.42
& 1376& 4.89& 0.39
\\
gpt-4.1-2025-04-14 &   20.71& 0.07&2.45&332& 2.38& 0.25& 879& 2.67& 0.27
& 657& 4.14& 0.23
\\
gemini-2.5-flash-lite &   30.38& 0.52&2.86&542& 4.12& 0.29& 1549& 3.95& 0.31& 773& 6.05& 0.34\\
\bottomrule 
\end{tabular}
\caption{Results of various structure supply chains in AIM-Bench under real-world metric settings.}
\label{tab:model_test}
\end{table*}

\begin{figure}[htbp]
    \centering
    
    \begin{minipage}[b]{0.49\linewidth}  
        \centering
        \includegraphics[width=0.99\linewidth]{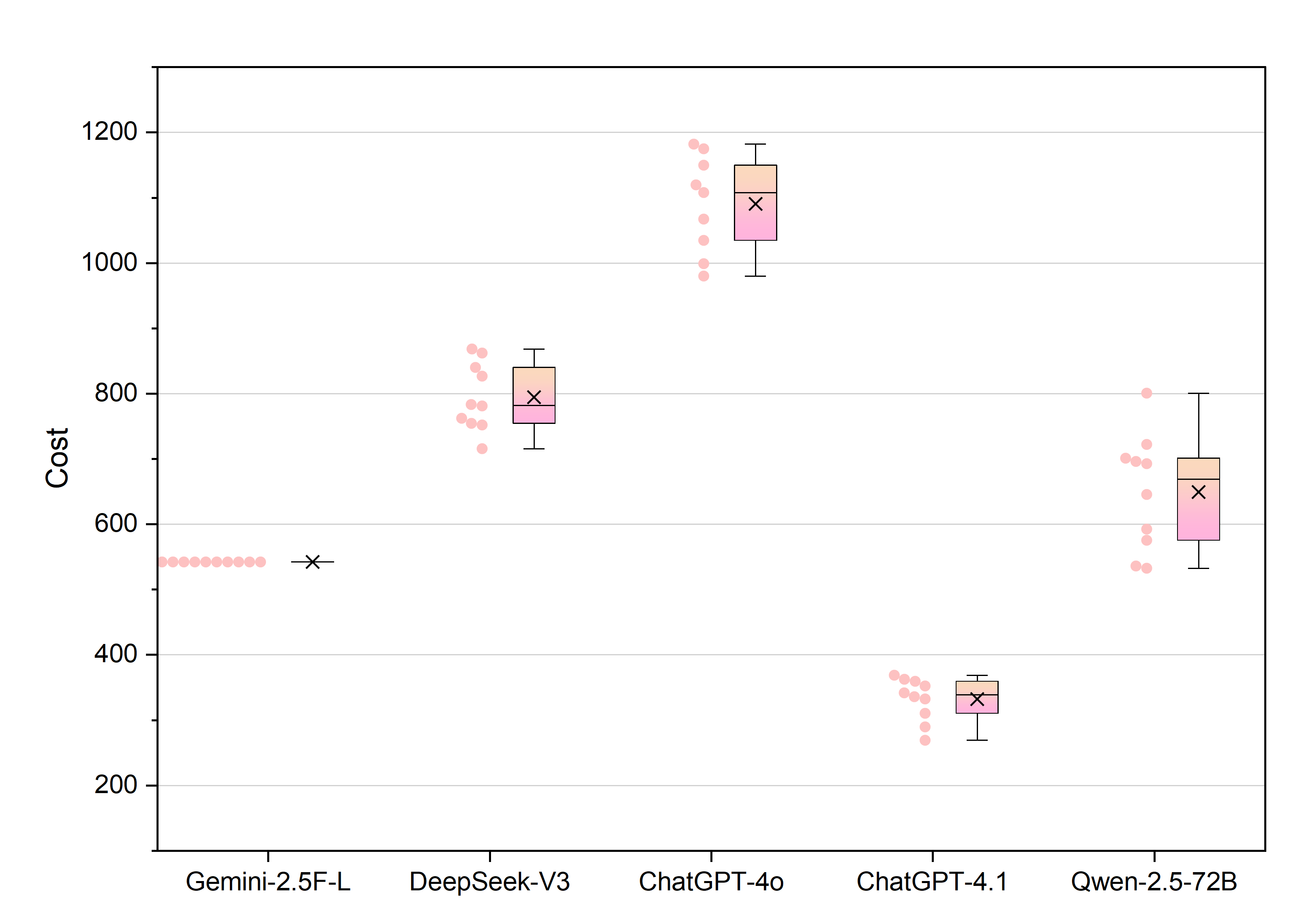}  
       
    \end{minipage}
    \hfill  
    \begin{minipage}[b]{0.49\linewidth}  
        \centering
        \includegraphics[width=0.99\linewidth]{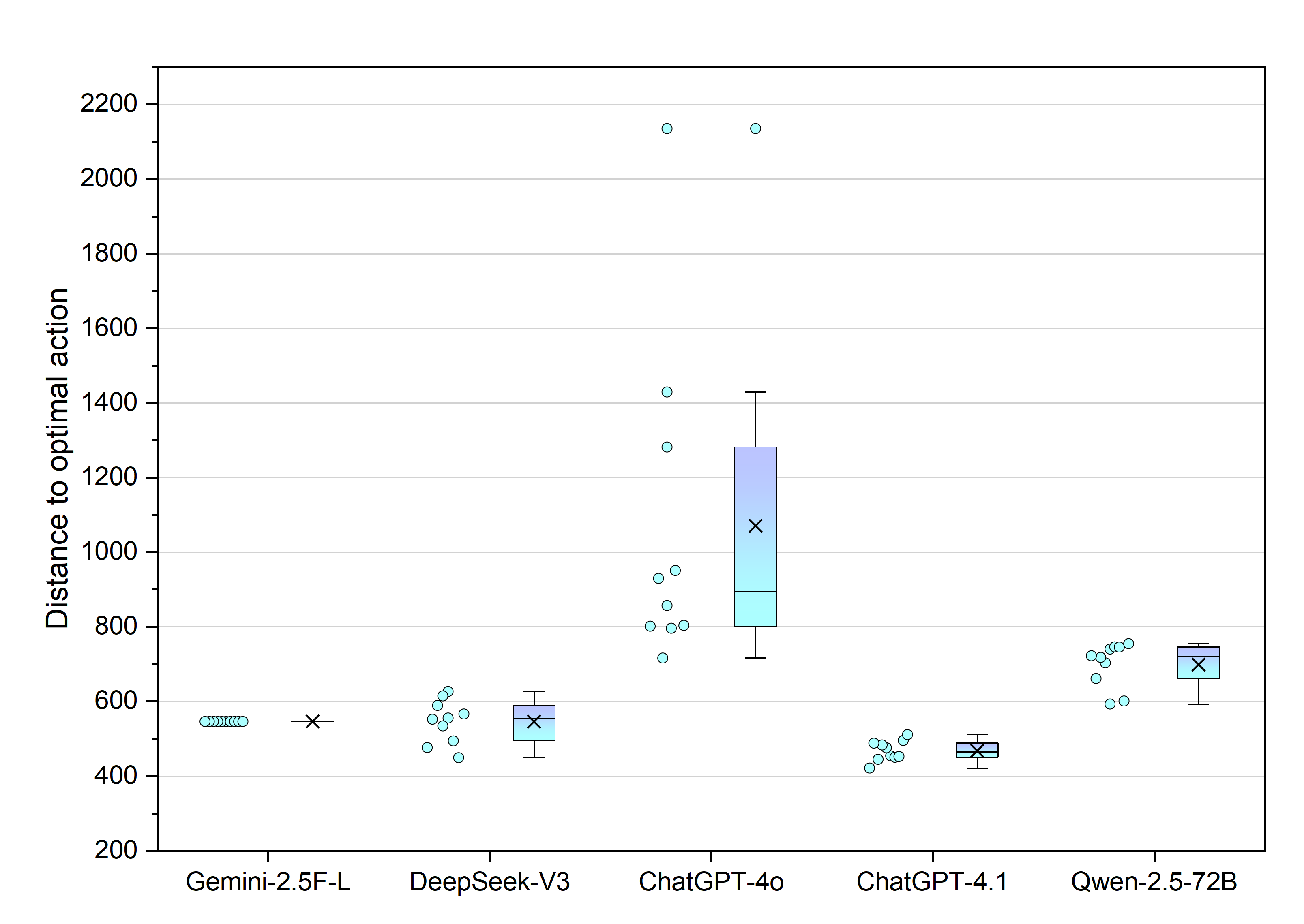}  
 
    \end{minipage}
    
    \caption{Inventory management cost and optimal replenishment distance results in AIM-Bench. Process metrics can contain more information.}  
    \label{fig:D}  
\end{figure}

\textbf{MPR-Finding V: The distance between LLMs order and the optimal order quantity is more informative and discriminating than the outcome metrics.} The results in multi-period replenishment decision are illustrated in Fig. \ref{fig:D} . Regarding the overall performance, the distance to the optimal order quantity serves as a more effective differentiator between models.For example, the GPT-4.1 model and the Qwen-2.5-72B model exhibit similar out-of-stock rates (0.250 and 0.256, respectively), but their distance metrics differ significantly: 467 for the GPT-4.1 model and 608 for the Qwen-2.5-72B model.This difference suggests that the GPT-4.1 parametric model performs better overall than the Qwen- 2.5-72B model. For models with large differences in turnover, stock-out rates and inventory management costs, for example, GPT-4.1 is 0.21  lower than the GPT-4o model in terms of stock-out rates and 758  ahead in terms of distance metrics, which suggests that there is a consistency in the difference in performance between models with significant differences.

\textbf{ALL-Finding VI: Individual models exhibit distinct characteristics under uncertainty.} For example, Qwen-2.5-72B exhibits a high inventory backlog with a maximum TR of 10.2  but a minimum SR of 0.2. As shown in Figure \ref{fig:TR-SR}. This suggests that Qwen-2.5-72B suffers from an over-ordering bias in response to demand uncertainty shocks relative to the optimal order quantity. Conversely, DeepSeek-V3 exhibits a lower TR (1.69) and a higher SR (0.54), while remaining closer to the optimal number of replenishments. This reflects the fact that DeepSeek exhibits less decision bias under uncertainty. On the other hand, the GPT-4o model exhibits better performance in the beer game with fixed lead time, however, it exhibits the highest inventory management cost in the results of the multi-cycle replenishment decision with random VLTs. Gemini-2.5-flash-lite's output shows surprising consistency, but again performs poorly in beer games. The results reveal potential limitations of the existing model, in particular, it is more difficult to cope with uncertainty in the timing of demand than in the magnitude of demand.

\section{Conclusions }

This study aimed to evaluate the decision-making capabilities and inherent biases of LLM agents in uncertain inventory management contexts, introducing the AIM-Bench benchmark for comprehensive LLM assessment. A primary finding reveals that LLM agents commonly exhibit human-like decision biases, such as mean anchoring and the bullwhip effect, though they demonstrate less susceptibility to demand chasing bias compared to human decision-makers. Furthermore, the research indicates that LLMs behavioral theories are context-dependent, as evidenced by the framing effect's inability to reliably alter their risk preferences. Crucially, the study validated effective mitigation strategies: cognitive reflection significantly reduced anchoring bias, while information sharing proved effective in ameliorating the bullwhip effect. These findings underscore the imperative for meticulous consideration of potential biases and the implementation of strategic model selection and bias mitigation when deploying LLMs in critical inventory decision-making scenarios. Ultimately, this research paves the way for the development of human-centered decision support systems capable of recognizing and counteracting AI's inherent irrationalities, thereby fostering more robust and reliable human-AI collaboration in complex supply chain challenges.

\section{Limitations}

A limitation of this work is that it only explores prompt-dependent methods to mitigate bias. We recommend that future work investigate RL-based approaches to improve decision support. In addition, the use of simulation environments simplifies the scenarios in which LLM is typically applied and may limit the generalisability of our findings.

The different levels of risk reversal observed in different contexts may stem from differences in training data and alignment strategies. However, the lack of publicly available information on specific training processes prevents us from drawing definitive conclusions. This highlights the need for community collaboration to study decision preferences in LLMs.

\bibliography{custom}

\end{document}